\documentclass[runningheads]{llncs}
\usepackage[T1]{fontenc}
\usepackage[utf8]{inputenc}
\usepackage[english]{babel}
\usepackage{xspace}
\usepackage{url}
\usepackage{hyperref}
\usepackage{xcolor}
\usepackage[utf8]{inputenc}
\usepackage{booktabs}
\usepackage[table]{xcolor}
\usepackage{multirow}
\usepackage{amsmath}
\usepackage{amsfonts}
\usepackage{enumitem}
\usepackage{listings}
\usepackage{multirow}
\usepackage{booktabs}
\usepackage{orcidlink}
\usepackage{tabularx}
\usepackage{array} 
\usepackage{cite}
\newcolumntype{Y}{>{\centering\arraybackslash}X}
\usepackage{graphicx}
\begin{document}
\title{Encoding Predictability and Legibility for Style-Conditioned Diffusion Policy}
%
%
\author{Adrien Jacquet Crétides \orcidlink{0009-0003-5666-6055} \and
Mouad Abrini \orcidlink{0009-0002-3727-4892} \and
Hamed Rahimi \orcidlink{0000-0001-9179-8625} \and
Mohamed Chetouani \orcidlink{0000-0002-2920-4539}} 
\authorrunning{A. Jacquet Crétides et al.}
%
\institute{Institut des Systèmes Intelligents et de Robotique (ISIR), CNRS UMR7222, Inserm ERL U1150, Sorbonne Université, Paris, France\\
\email{\{first\_name.last\_name\}@sorbonne-universite.fr}\\ 
}
\maketitle              
\begin{abstract}
Striking a balance between efficiency and transparent motion is a core challenge in human-robot collaboration, as highly expressive movements often incur unnecessary time and energy costs. In collaborative environments, legibility allows a human observer a better understanding of the robot's actions, increasing safety and trust. However, these behaviors result in sub-optimal and exaggerated trajectories that are redundant in low-ambiguity scenarios where the robot's goal is already obvious. To address this trade-off, we propose Style-Conditioned Diffusion Policy (SCDP), a modular framework that constrains the trajectory generation of a pre-trained diffusion model toward either legibility or efficiency based on the environment's configuration. Our method utilizes a post-training pipeline that freezes the base policy and trains a lightweight scene encoder and conditioning predictor to modulate the diffusion process. At inference time, an ambiguity detection module activates the appropriate conditioning, prioritizing expressive motion only for ambiguous goals and reverting to efficient paths otherwise. We evaluate SCDP on manipulation and navigation tasks, and results show that it enhances legibility in ambiguous settings while preserving optimal efficiency when legibility is unnecessary, all without retraining the base policy. 

\end{abstract}

\begin{figure}[ht]
    \centering
    \includegraphics[width=1.0\linewidth]{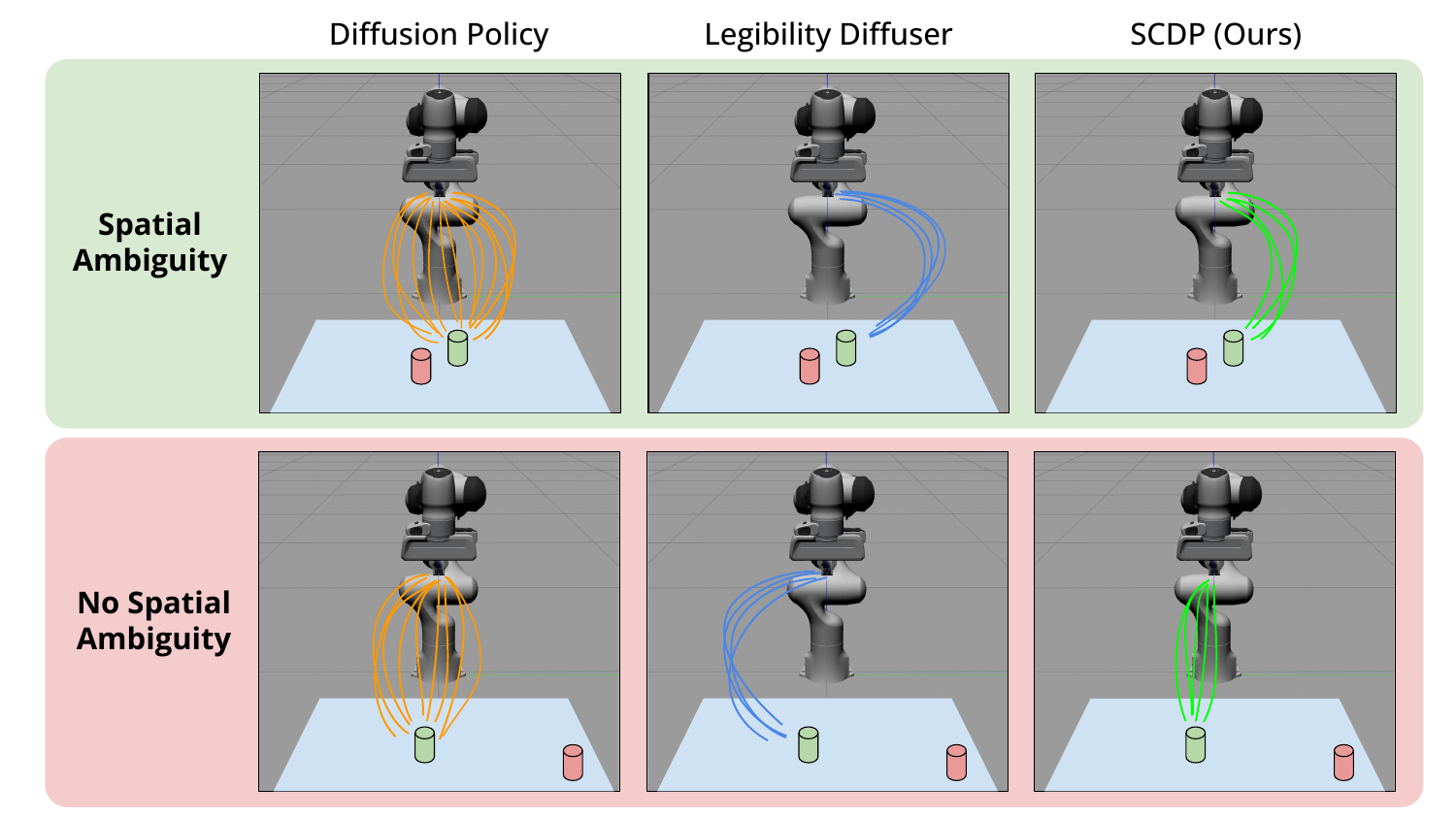}
    \caption{Style-Conditioned Diffusion Policy is an offline imitation learning framework that allows for motion conditioning depending on the environment’s context. In ambiguous scenes (top), SCDP produces intent-expressive motion to resolve goal ambiguity. When ambiguity is low (bottom), it prioritizes task efficiency, avoiding sub-optimal and exaggerated trajectories.}
    \label{fig:main_fig}
\end{figure} 

\section{Introduction}

Achieving a balance between operational efficiency and transparent robot motion remains a significant challenge in human-robot collaboration \cite{LegPred}. While these expressive behaviors often result in sub-optimal or exaggerated trajectories, they are a necessary trade-off to ensure motion legibility, a modality where the robot’s movement alone allows an external observer to correctly infer its intended goal \cite{generating_legible_motion}. This becomes especially critical on minimalist platforms or in constrained environments where other communication channels, such as speech or gaze \cite{Wallkotter2021Explainable} are unavailable. Ultimately, legibility is essential to achieve safe and intuitive interactions between humans and robots \cite{dragan2015}. To acquire such expressive behaviors, Imitation Learning has recently served as a foundational approach, enabling robots to mimic strategies directly from data \cite{zare2024}. Building upon this, Diffusion Models \cite{DDPM} have emerged as a solution offering a robust framework for modeling these complex trajectory distributions in a stochastic yet controllable manner \cite{survey_diffusion}. \\

While legibility is an implicit but powerful form of communication, it often comes at the cost of efficiency. Making a trajectory more expressive may require deviations from a classical path, that would be more direct or optimal from an usual robot generation point of view \cite{generating_legible_motion}. In many practical scenarios, being legible is not always necessary \cite{LegPred}. For example, when the goal is clearly distinguishable or when ambiguity is low, a standard, efficient trajectory can be sufficient for an external observer to infer quickly what is the goal of the motion. Therefore, finding a trade-off between legibility and efficiency is a key challenge, by developing systems that can adaptively produce legible motion only when required, balancing expressiveness with task performance. \\

We introduce in this work \textit{Style-Conditioned Diffusion Policy} (SCDP), a novel approach that modulates trajectory generation by constraining a diffusion model toward a specific style. Depending on the spatial ambiguity of the environment, SCDP dynamically shifts the robot’s motion between legibility and predictability, as shown in figure \ref{fig:main_fig}.
Our method consists of a post-training pipeline, freezing the base diffusion model and training external environment-aware modules, namely a scene encoder and legibility/predictability predictor multi-layer perceptrons, to generate an additional conditioning based on the scene ambiguity without requiring any modification to the diffusion model's internal structure. \\

Through experiments, we show that our architecture: (i) enables style-conditioned trajectory generation without retraining or altering the base policy, (ii) supports adaptive and scene-dependent modulation based on an ambiguity detection mechanism and (iii) improves motion interpretability in ambiguous scenarios while focusing on efficiency when legibility is not needed.

\section{Related Work}

\subsection{Legible Motion Generation}

The concept of legible motion for robotics has emerged as a key-topic for HRI, emphasizing the design of movements that clearly communicates the robot's intentions to human observers \cite{Lichtenthler2011TowardsAL}. Dragan et al. \cite{LegPred} first introduced the distinction between legible and predictable motion, and modeled human inference through a Bayesian framework, relying on cost functions and optimization to maximize the probability of the correct goal being identified by the human observer, given the robot's trajectory from its start to its current position \cite{generating_legible_motion}. Learning-based approaches have since gained prominence, utilizing reinforcement learning to optimize legibility metrics \cite{bied2020} and supervised observer models to predict human goal inference \cite{slotv}. A notable example is Legibility Diffuser \cite{LegibilityDiffuser}, which employs diffusion with classifier-free guidance \cite{classifer_free} to generate legible motion. However, this method remains agnostic to environment's spatial ambiguity and relies on manually tuned guidance weights at test time, which motivates the need for a more environment-aware and interpretable mechanism to encode legibility into the generation process. 

\subsection{Robot Motion and Conditional Generative Models}

Generative models have gained increasing attention in robotic motion generation due to their ability to learn complex, high-dimensional distributions of motion data from demonstrations \cite{urain2024deepgenerativemodelsrobotics}. More recently, diffusion models have shown promising results in generating diverse and temporally coherent robot trajectories, especially in manipulation and locomotion tasks \cite{survey_diffusion}. Notably, Diffusion Policy \cite{Diffusion_Policy} still demonstrates competitive performance across visuomotor manipulation tasks. Generation is conditioned on observation and current robot's state, using the FiLM method \cite{film}. 
Beyond the scope of robotics, recent advancements in generative modeling have introduced conditioning techniques to constrain the diffusion process \cite{berrada2024on, cond_diff_survey}. Specifically, Li et al \cite{c-vector} added an additional context vector to constrain the diffusion generation, allowing for fine-grained control over the sampling process without altering the base model. Inspired by this, we propose an external, environment-aware module that modulates trajectory generation based on the scene's context.

\section{Preliminaries}

\subsection{Diffusion Policy}

Our work builds upon Diffusion Policy in its U-Net form \cite{Diffusion_Policy, unet}, which adapts the Denoising Diffusion Probabilistic Models (DDPM) \cite{DDPM} framework to learn conditioned action sequences from demonstrations. The denoising update rule is defined as:

\begin{equation}
    X_{t}^{k-1} = \alpha (X_{t}^{k} - \gamma \epsilon_\theta (O_t, X_{t}^{k}, k) + \mathcal{N}(0, \sigma^2I))
\end{equation}

where $X_t^k$ represents the noisy action at step $k$, and $\epsilon_\theta$ estimates the noise component added to the action sequence, conditioned on the observation $O_t$. While observation can be vision-based, we choose to limit it in our work to the goal state $g^*$ and the current state of the robot $s_t$.

\subsection{Legibility and Predictability}

Legibility refers to the robot's ability to communicate its intended goal to a human observer through the expressive characteristics of its motion alone \cite{generating_legible_motion}. By doing legible, or intent expressive actions, a robot allows an external observer to have a better understanding of the robot's action, increasing safety and efficiency in human-robot collaboration \cite{dragan2015, leg_safety}. A legible motion will seek to maximize the probability that an external observer can infer which objective is targeted, even if it means reducing the effectiveness of the motion, by achieving a longer and sub-optimal trajectory. It can be defined as follows \cite{generating_legible_motion}:

\begin{equation}
    \text{legibility}(\xi) = 
\frac{
    \int P(g^* \mid \xi_{s \rightarrow \xi(t)}) f(t) \, dt
}{
    \int f(t) \, dt
}
\end{equation}

where $g^*$ is the targeted goal, $\xi(t)$ represents the partial trajectory up to time $t$, and $f(t)$ is a temporal weighting function (e.g., $f(t) = t$), emphasizing early steps of the trajectory, where observer’s uncertainty is the highest due to multiple potential goals being equally plausible. 

While legibility focuses on the inference of the goal from the motion, predictability reflects the observer's ability to anticipate the trajectory itself, provided the goal is already known. Mathematically, if legibility maximizes $P(g \mid \xi)$, predictability seeks to maximize $P(\xi \mid g)$. A predictable trajectory aligns with the human observer's expectation of how an action should be performed, which typically corresponds to the most efficient or cost-optimal path. 

\subsection{Spatial Ambiguity} \label{subsub:ambiguity}

We focus in this work on spatial ambiguity, which depends on the environment's configuration. Given a goal space $G$, we formalize a scene as spatially ambiguous for a target goal $g^* \in G$ relative to a set of distractor goals $g^-$ by examining the observer's ability to infer the correct intent from an efficient trajectory. Let $\xi^*$ denote an optimal trajectory toward $g^*$, typically characterized by the shortest Euclidean distance. 
Formally, given a confidence threshold $\tau$, we define a scene as ambiguous if there exists at least one goal  $g^- \in G \setminus \{g^*\}$ such that the following condition holds for the efficient trajectory $\xi^*$:

\begin{equation}
    P(g^* | \xi_{s \rightarrow \xi(t)}^*) - P(g^- | \xi_{s \rightarrow \xi(t)}^*) < \tau
\end{equation}

Where $P(g | \xi_{s \rightarrow \xi(t)})$ represents the posterior probability of a goal $g$ given the observed partial trajectory $\xi^*_{s \rightarrow \xi(t)}$.

\section{Style-Conditioned Diffusion Policy}

We propose a composite framework that builds on Diffusion Policy \cite{Diffusion_Policy} as the base model and uses an additional module to condition it. It is composed of three Multi-Layer Perceptrons (MLPs), one acting as a scene encoder and generating a context vector corresponding on the environment the robot acts in, and the two others acting as conditioning predictors depending on this context. At inference time, we use an ambiguity detection module to determine if legibility or efficiency is required.

\subsection{Task Formalization}

We consider robot motion generation as a sequential decision-making problem in which an agent must move toward a specific goal.
Let $G \subset \mathbb{R}^2$ denote the set of possible goals, where $g \in G$ represents a position for the robot to reach. We model this problem as a discrete-time, infinite-horizon Markov Decision Process (MDP),
$\mathcal{M} = (\mathcal{S}, \mathcal{A}, \mathcal{T}, \rho_0)$,
where $\mathcal{S}$ is the state space, $\mathcal{A}$ the action space, $\mathcal{T}(\cdot | s, a)$ the state transition distribution, and $\rho_0(\cdot)$ the initial state distribution.
At each step, the agent observes a state $s_t$ and uses a policy $\pi$ to select an action $a_t = \pi(s_t)$, where the states and actions correspond to the robot’s current and next joint configurations. The agent then transitions to the next state $s_{t+1} \sim \mathcal{T}(\cdot | s_t, a_t)$.
We assume having a dataset of $N_d$ task demonstrations
$D = \{\xi_i\}_{i=1}^{N_d}$
where each demonstration is a trajectory
$\xi_i = (s_{i0}, a_{i0}, s_{i1}, a_{i1}, \ldots, s_{iT})$, along with the environment state, namely a positive goal $g^* \in G$.
In the case of legible motion generation, our objective is to generate motion toward $g^*$ that is distinct from motion toward any negative goal $g^- \in G \setminus \{g^*\}$, such that
$p(g^* \mid s_t, a_t) > p(g^- \mid s_t, a_t)$.
On the other hand, in the case of efficient motion, our objective is to minimize the cost of reaching $g^*$, which corresponds to maximizing the likelihood of the trajectory given the goal, $p(\xi \mid g^*)$.

\subsection{Scene Encoding}

As legibility is entirely depending on the environment's configuration and the present objects' positions, we introduce a scene encoder, that is trained independently from the rest of the architecture, to learn a latent representation of the environment and of the spatial relations between the different goals. This MLP takes as inputs the different goals' coordinates $g^*$ and $g^-$ in the scene. For every goal $g^-_i \in G \setminus \{g^*\}$, we compute \( r_i = g^-_i - g^*\) the relative vector to $g^*$, and \( j_i = \|g^-_i - g^*\|_2 \in \mathbb{R} \) the euclidean distance to $g*$. We then construct the enriched vectors $\tilde{g}_i = \begin{bmatrix} g^-_i & r_i & j_i \end{bmatrix}^\top$.
The $g^*$ enriched vector is defined as $\tilde{g^*} = \begin{bmatrix} g^* & \mathbf{0} & 0 \end{bmatrix}^\top \in \mathbb{R}^5$, acting as the origin of the scene's coordinate system and grounding the relative vectors of the other goals.
Given $N$ the number of negative goals, we can concatenate these enriched vectors into a single one, $ x = \begin{bmatrix} \tilde{g^*}^\top & \tilde{g}_1^\top & \dots & \tilde{g}_N^\top \end{bmatrix}^\top \in \mathbb{R}^{5\times(N+1)} $, that will be passed into the encoder \( S : \mathbb{R}^{5\times(N+1)} \to \mathbb{R}^{s} \) to get a latent contextual vector of the scene :

\begin{equation}
    c = S(x) \in \mathbb{R}^s
\end{equation}

To train this encoder, we employ a reconstruction-based approach using an autoencoder architecture \cite{autoencoder1986}, where the scene encoder is jointly trained with a decoder.

\subsection{Constraining the Diffusion}

\begin{figure}[ht]
    \centering
    \begin{minipage}[b]{0.48\textwidth}
        \centering
        \includegraphics[height=1.65in]{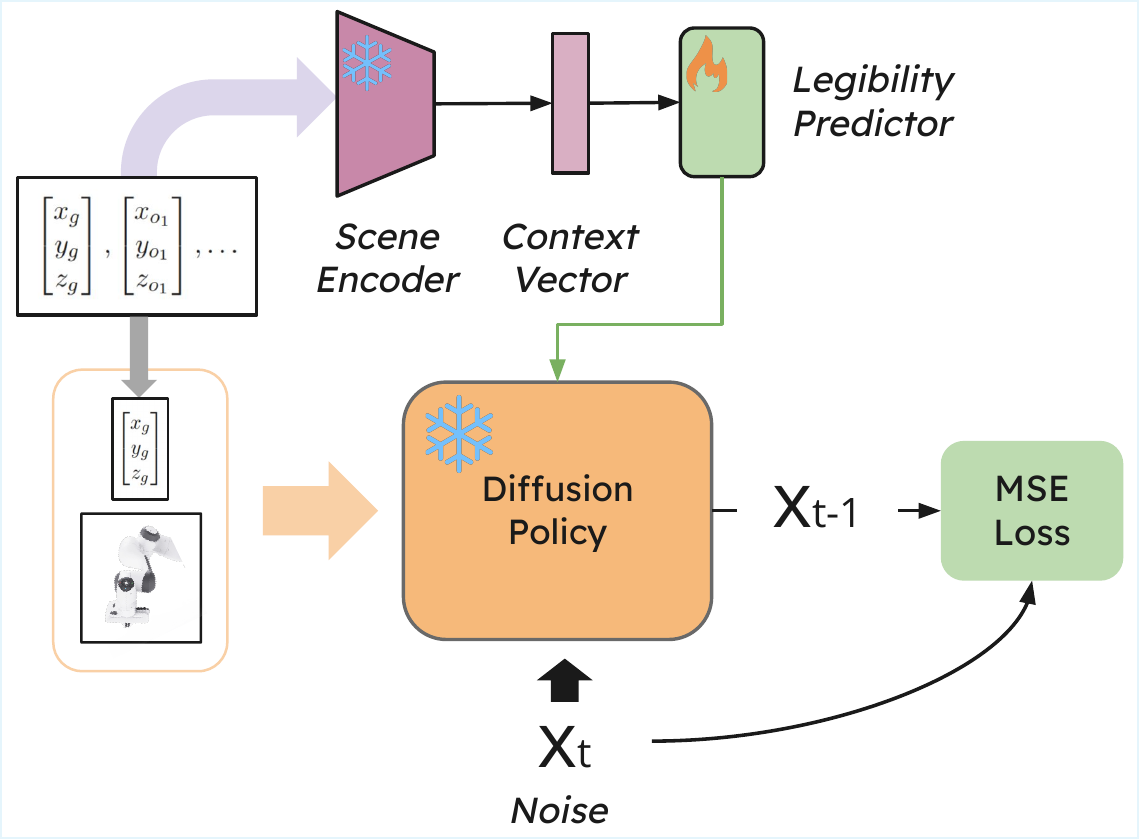}
        \vfill
        {\small (a) Training Process.}
    \end{minipage}
    \hfill
    \begin{minipage}[b]{0.48\textwidth}
        \centering
        \includegraphics[height=1.65in]{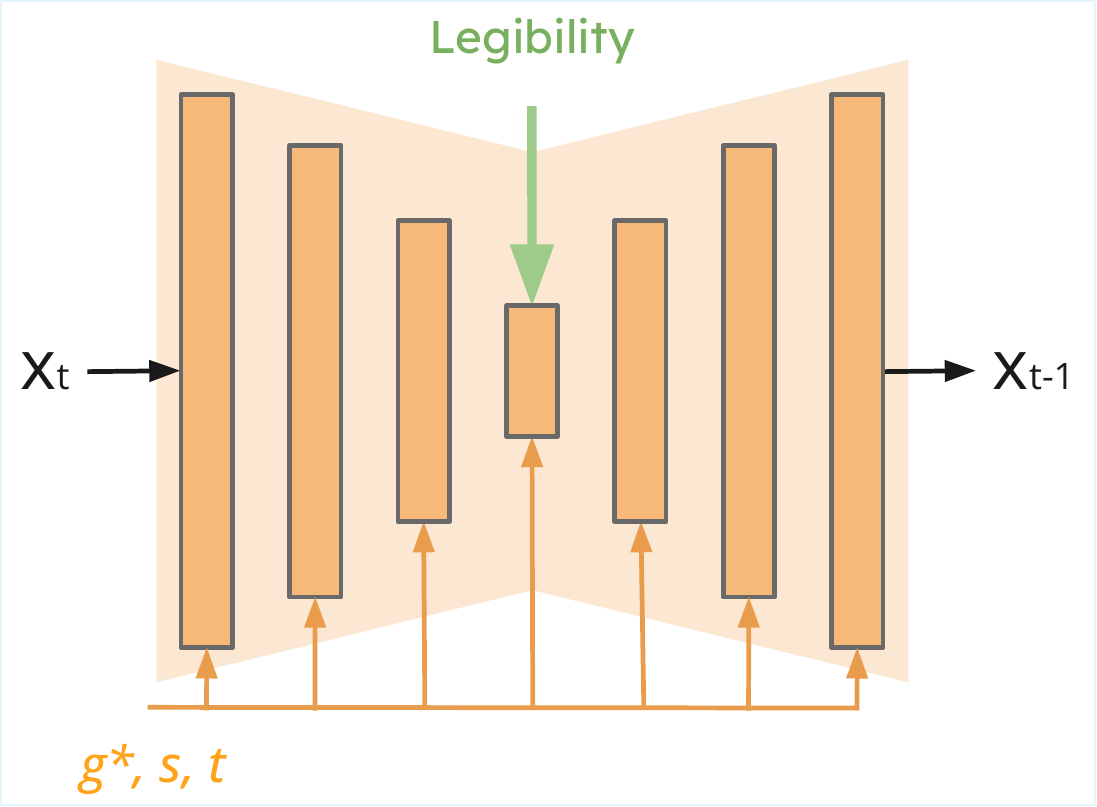}
        \vfill
        {\small (b) U-Net Conditioning.}
    \end{minipage}

    \caption{(a) The predictor module is integrated via a post-training pipeline where the base Diffusion Policy weights remain frozen. By training the lightweight MLP on a subset of expressive demonstrations, the module learns to specifically compensate for the residuals between the style-specific trajectories and the general paths the base model was originally trained to reproduce.(b) The conditioning from the predictor is only applied to the bottleneck of the diffusion U-Net using FiLM to denoise $X$ at each timestep $t$.}
    \label{fig:Architecture}
\end{figure}

We design a second MLP, responsible of the trajectory style encoding. Figure \ref{fig:Architecture} shows the post training process. After training the base Diffusion Policy's U-Net on a large set of demonstrations, we freeze its weights and add the scene encoder, also pretrained, and the predictor MLP. We then train this new composite model on a smaller dataset, containing only style-specific demonstrations.
As the base model is frozen, only the style MLP is updated by the MSE loss through this training phase, and has to learn to compensate through the generation the difference between the legible demonstrations seen here and the classical demonstrations Diffusion Policy has been trained to reproduce. Training on a specific data subset, the context vector $c$ generated by the scene encoder passes through the MLP, that generates two vectors $\gamma$ and $\beta$:

\begin{equation}
    \gamma = W_\gamma c + b_\gamma \in \mathbb{R}^{l}, \quad
\beta = W_\beta c + b_\beta \in \mathbb{R}^{l}
\end{equation}

where $l$ is the feature dimension of the bottleneck layer, \( W_\gamma, W_\beta \in \mathbb{R}^{l \times d} \) and \( b_\gamma, b_\beta \in \mathbb{R}^{l} \) are learned weight matrices. These parameters \( \gamma, \beta \) are then used to modulate the U-Net's middle layer vector $h$ using a FiLM conditioning \cite{film}:

\begin{equation}
    \text{FiLM}(h) = \gamma \odot h + \beta, \quad h \in \mathbb{R}^{l}
\end{equation}

This results in an additional module that can be added to the base diffusion model to constrain its generation to a desired controlled distribution.

\subsection{Ambiguity Detection}

\begin{figure}[ht]
    \centering
    \begin{minipage}[b]{0.48\textwidth}
        \centering
        \includegraphics[height=1.65in]{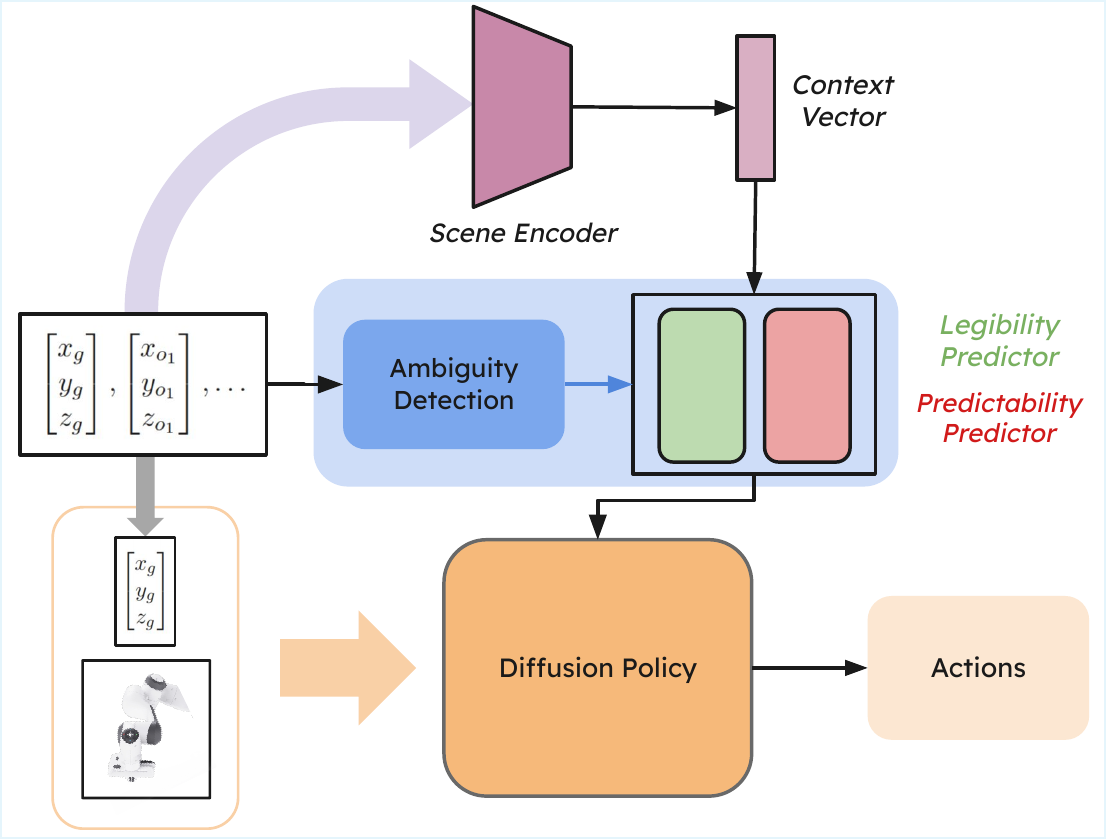}
        \vfill
        {\small (a) Evaluation Process.}
    \end{minipage}
    \hfill
    \begin{minipage}[b]{0.48\textwidth}
        \centering
        \includegraphics[height=1.65in]{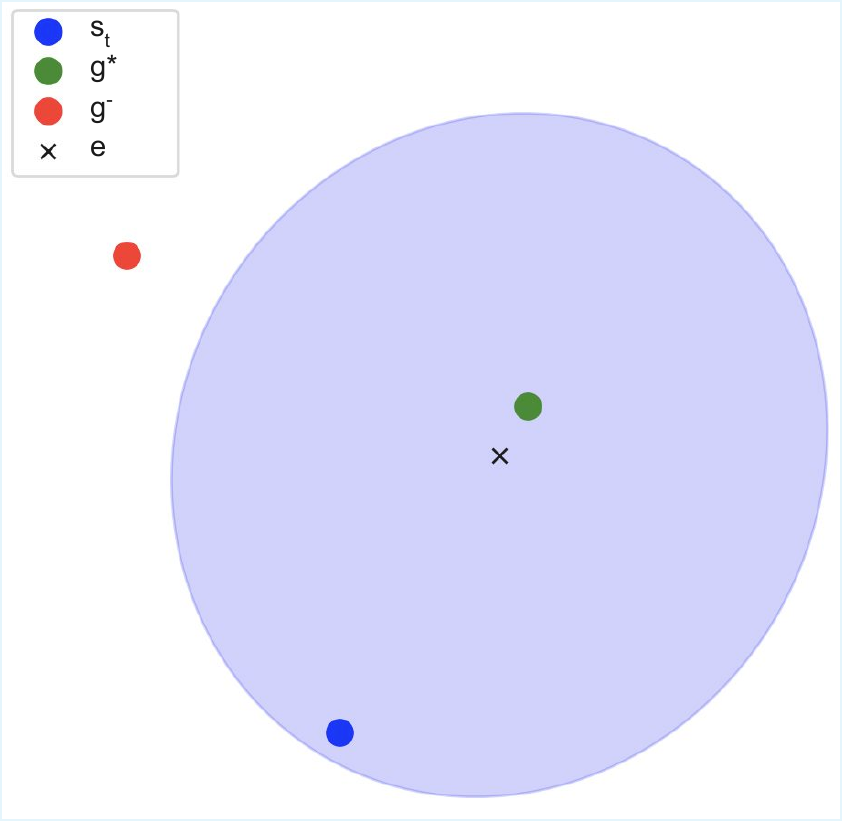}
        \vfill
        {\small (b) Ellipse of Ambiguity.}
    \end{minipage}

    \caption{(a) The environment state is passed through the ambiguity detection module to determine if the scene is spatially ambiguous and decide which conditioning should be used. (b) Visualization of the ellipse of ambiguity used for the scene's classification. The scene is labeled as spatially ambiguous when $g^-$ falls inside the elliptical boundary.}
    \label{fig:inference}
\end{figure}

To approximate the probabilistic definition of spatial ambiguity from \ref{subsub:ambiguity}, We define an ellipse of ambiguity $\mathcal{E}$. This serves as a lightweight geometrical proxy representing the spatial zone of confusion where a negative goal $g^-$ is most likely to cause an observer to misinterpret the robot's intent. The ellipse is constructed such that the robot's current state $s_t$ acts as one of its focal points, and is centered at a point $e$ located between the robot's current state $s_t$ and the goal $g^*$. This center is defined as: $e = s_t + \kappa(g^* - s_t)$, where $\kappa \in (0.5, 1)$ is a scaling factor. A scene is considered spatially ambiguous if any negative goal $g^-$ falls within this elliptical boundary:

\begin{equation}
    \text{Ambiguity}(g^*, g^-, s_t) = \begin{cases} 1 & \text{if } (g^- - e)^\top M (g^- - e) \le 1 \\ 0 & \text{otherwise} \end{cases}
\end{equation}

Here, $M$ is a symmetric positive-definite matrix that defines the orientation and the semi-axes of the ellipse.
Figure \ref{fig:inference} shows the architecture used at inference time and a visualization of the ellipse of ambiguity for a given environment. The ambiguity detection module serves as an arbitrator that selects the optimal conditioning for the Diffusion Policy based on the environment's risk of goal-confusion. If the scene is flagged as spatially ambiguous, the legibility predictor is activated to constrain the diffusion process toward a more expressive trajectory. Otherwise, the predictability predictor constrains the generation to the most efficient path. While simplified, this proxy allows for real-time arbitration with negligible computational overhead, maintaining the modularity of the SCDP framework. 

\section{Evaluation}

Our experiments are designed to evaluate how effectively the model balances legibility and efficiency across varying environmental contexts. To address this trade-off, we consider the following research questions:
\begin{itemize}
    \item[$\bullet$] Can SCDP increase legibility performance when the situation requires higher clarity?
    \item[$\bullet$] Does SCDP effectively reduce legibility to prioritize efficiency in unambiguous environments?
\end{itemize}
To provide robust answers, we test these questions across diverse datasets featuring a large range of spatial configurations.

\subsection{Datasets}

We conduct our experiments on two tasks. Each task is divided into two environmental scenarios: spatial ambiguity and no spatial ambiguity.

\textbf{Block Reach}: the Block Reach task is a benchmark commonly used in legibility studies \cite{LegPred}. In this setup, two objects are placed randomly on the scene, and a manipulator arm, here a Franka Emika Panda robot, is required to reach one of them. 

\textbf{Navigation}: In this task, a mobile robot has to reach one of the two goals present in the room. The robot used here is a Turtlebot. 

To train our agents, we collected datasets of 200 demonstrations for each task using the Gazebo simulator to perform and record the robots' states and actions. These demonstrations were procedurally generated using quadratic Bezier curves to ensure a diverse distribution of trajectories that range from near-optimal straight lines to highly curved paths.

\subsection{Baselines}

\begin{figure}[tbp]
    \includegraphics[width=\textwidth]{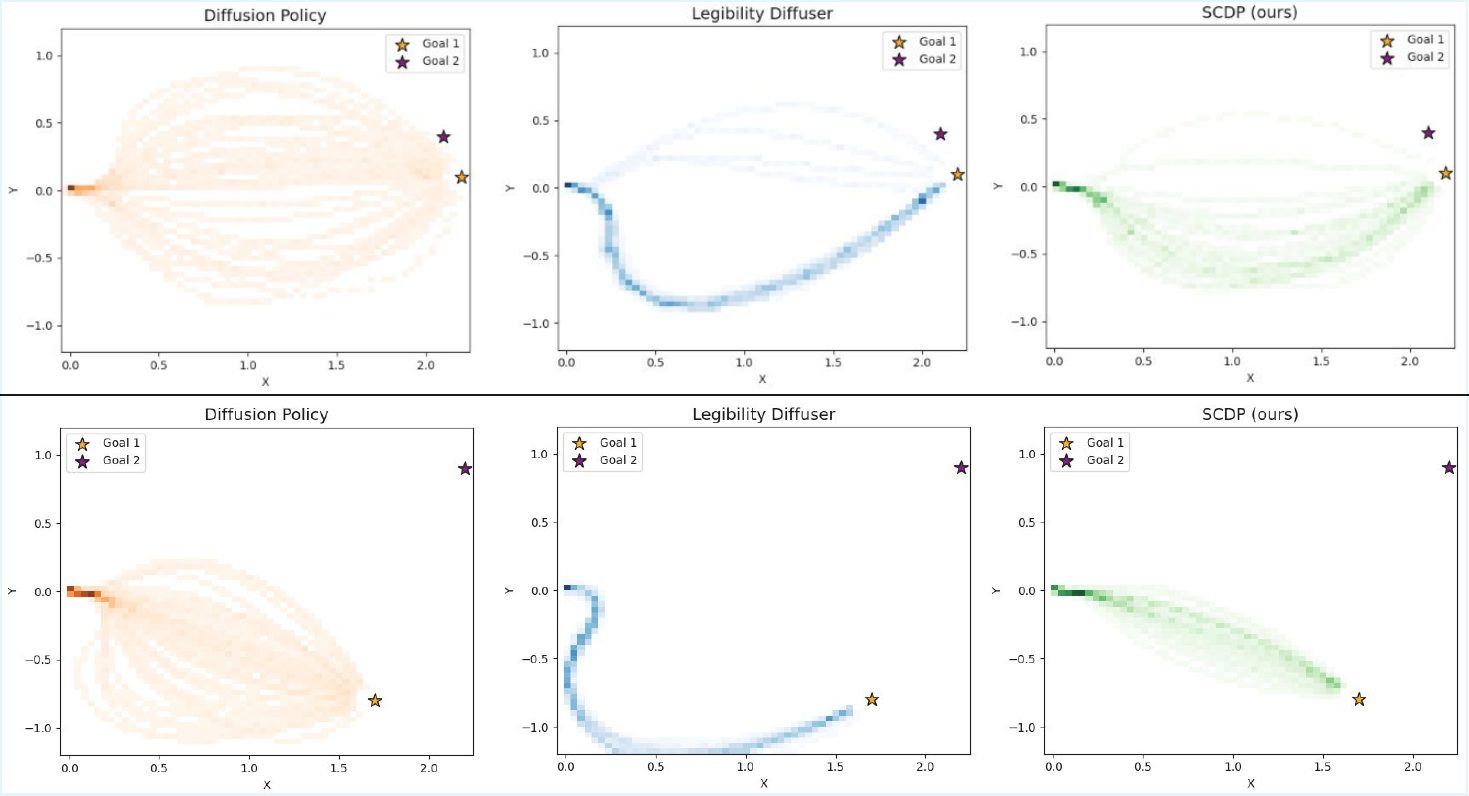}
    \caption{Visualization of SCDP and baselines' inferences in ambiguous (top) and non-ambiguous (bottom) scene configurations for the navigation task. While Diffusion Policy captures the entire data distribution and Legibility Diffuser collapses on the most legible mode, SCDP constrains its generation depending on the scene configuration.} \label{fig:example}
\end{figure} 

We compare and evaluate our method with respect to other diffusion-based baselines.

\textbf{Diffusion Policy \cite{Diffusion_Policy}}: Diffusion Policy is the original DDPM-based model our architecture is built on. Similarly to our method, it is here used in its goal-conditioned form, to isolate the effect of our contribution. 

\textbf{Legibility Diffuser \cite{LegibilityDiffuser}}: Also built on a Diffusion Policy, Legibility Diffuser is a variant that introduces a guidance term to pull the trajectory away from a negative goal $g^-$ while going to $g^*$. As the official implementation is not available, their method was reproduced for comparison, using the following values for their hyperparameters: $w_t = 4.5$, $\alpha = 0.9$ and $\lambda = 0.5$. Parameters were chosen to recreate a similar behavior to the one presented in the evaluation of the baseline.

\textbf{Training Dataset}: We additionally report the values of the datasets used to train the different baselines.

\subsection{Implementation Details}

This section details the architectural and training configurations for the SCDP framework and the comparative baselines.

\textbf{Base Diffusion Policy:} We employ a U-Net-based Diffusion Policy \cite{Diffusion_Policy} with approximately $70$M parameters for the different baselines and SCDP. The base model is trained on a foundational dataset of $200$ expert demonstrations for $100$ epochs.

\textbf{Scene Encoder:} To extract latent environmental context, we utilize a 3-layer MLP architecture comprising $26,000$ parameters. This module is pre-trained using a reconstruction loss on $5,000$ randomized goal configurations over $50$ epochs to ensure a robust representation of spatial relations.

\textbf{Conditioning Predictors:} The style-specific modulation is handled by two separate 4-layer MLP backbones ($1$M parameters each) representing the legibility and predictability predictors. These modules are fine-tuned for $300$ epochs using specialized subsets containing the top $20\%$ most legible and $20\%$ most efficient demonstrations from the original datasets, respectively.

\subsection{Metrics}

We evaluate the generated trajectories using two primary performance indicators, which are then combined into a final adaptive transparency score metric.

\subsubsection{Primary Metrics}

The first metric, detachment score, was introduced by the authors of Legibility Diffuser \cite{LegibilityDiffuser}, and serves as a proxy of legibility when targeted and negative goals are close. It quantifies the divergence from the negative goal $g^-$ along the trajectory $\xi_{s \rightarrow g^*}$:

\begin{equation}
    D(\xi_{s_0 \rightarrow g^*}) = \sum_{s_t \in \xi_{s \rightarrow g^*}} \frac{\|g^{-} - s_t\|_2}{t}
\end{equation}

The second metric is trajectory efficiency, calculated as the reciprocal of the total Euclidean distance traveled during motion execution:

\begin{equation}
    E(\xi_{s_0 \rightarrow g^*}) = \left( \sum_{s_t \in \xi_{s \rightarrow g^*}} \left\| s_t - s_{t-1} \right\|_2 + \epsilon \right)^{-1}
\end{equation}

where $\epsilon$ is a small constant to ensure numerical stability. 
Values for $D$ and $E$ are normalized using the Min-Max method with respect to the values obtained from the training dataset to ensure they share a comparable scale.

\subsubsection{Adaptive Transparency Score}

To evaluate how effectively the model navigates the conflict between legibility and efficiency, we define a trade-off metric, the adaptive transparency score ($T$):
\begin{equation}
T = (1 -w_{amb})\hat{D} + w_{amb}\hat{E}
\end{equation}

where $\hat{D}$ and $\hat{E}$ are the normalized detachment and efficiency scores. The weight $w_{amb}$ is modeled as a continuous sigmoid function of the Euclidean distance $j$ between the goals $g^*$ and $g^-$, which is also normalized using the Min-Max method:

\begin{equation}
w_{amb}(j) = \frac{1}{1 + e^{-u(j - x_0)}}
\end{equation}

This formulation allows for a fluid transition between behavioral modes based on the environment's spatial configuration, quantifying the agent’s ability to adapt its motion style to the specific geometric requirements of the scene.
In our evaluation, we set the steepness parameter $u=2.5$ to ensure a fluid yet distinct transition between behavioral modes, and the midpoint distance $x_0=0.5$ as it is the center of the normalized distance domain.

\subsection{Results} 

\begin{table}[tbp]
\centering
\caption{Performance Comparison using Adaptive Transparency Score for the Block Reach and Navigation tasks.}
\label{tab:fused_results}
\small
\begin{tabularx}{\textwidth}{l YY} 
\toprule
\textbf{Method} & \textbf{Spatial Ambiguity} & \textbf{No Spatial Ambiguity} \\
\midrule

\rowcolor{gray!15} \multicolumn{3}{l}{\textit{Block Reach Evaluation}} \\
\addlinespace 
Diffusion Policy    & $0.52 \pm 0.06$ & $0.61 \pm 0.06$ \\
Legibility Diffuser & $\mathbf{0.61 \pm 0.09}$ & $0.43 \pm 0.10$ \\
Ours (SCDP)         & $\mathbf{0.58 \pm 0.08}$ & $\mathbf{0.74 \pm 0.06}$ \\
\cmidrule(lr){1-1}
Dataset             & $0.47 \pm 0.17$ & $0.62 \pm 0.18$ \\

\addlinespace 
\rowcolor{gray!15} \multicolumn{3}{l}{\textit{Navigation Evaluation}} \\
\addlinespace 
Diffusion Policy    & $0.50 \pm 0.03$ & $0.61 \pm 0.05$ \\
Legibility Diffuser & $\mathbf{0.62 \pm 0.02}$ & $0.32 \pm 0.12$ \\
Ours (SCDP)         & $\mathbf{0.59 \pm 0.03}$ & $\mathbf{0.76 \pm 0.06}$ \\
\cmidrule(lr){1-1}
Dataset             & $0.46 \pm 0.17$ & $0.59 \pm 0.20$ \\
\bottomrule
\end{tabularx}
\end{table}

Table \ref{tab:fused_results} presents the evaluation results using the adaptive transparency score $T = (1-w_{amb})\hat{D} + w_{amb}\hat{E}$ across different methods, while table \ref{tab:separated_results} displays the detachment and trajectory efficiency scores separately.
Results are averaged in simulation over 100 inferences. Figure \ref{fig:example} illustrates the qualitative differences in trajectory generation between the baselines in both ambiguous and non-ambiguous settings. 

\begin{table}[tbp]
\centering
\caption{Separated Detachment and Trajectory Efficiency Scores Comparison for Block Reach and Navigation tasks.}
\label{tab:separated_results}
\small
\newcolumntype{Y}{>{\centering\arraybackslash}X} 
\begin{tabularx}{\textwidth}{l YYYY} 
\toprule
\multirow{2}{*}{\textbf{Method}} & \multicolumn{2}{c}{\textbf{Spatial Ambiguity}} & \multicolumn{2}{c}{\textbf{No Spatial Ambiguity}} \\
\cmidrule(lr){2-3} \cmidrule(lr){4-5} 
 & \textbf{Detachment} & \textbf{Efficiency} & \textbf{Detachment} & \textbf{Efficiency} \\
\midrule
\rowcolor{gray!15} \multicolumn{5}{l}{\textit{Block Reach Evaluation}} \\
\addlinespace 
Diffusion Policy    & $0.52 \pm 0.06$ & $0.67 \pm 0.06$ & $0.43 \pm 0.09$ & $0.65 \pm 0.09$ \\
Legibility Diffuser & $0.89 \pm 0.12 \color{green!60!black}{\uparrow}$ & $0.13 \pm 0.10 \color{red}{\downarrow}$ & $0.67 \pm 0.16 \color{green!60!black}{\uparrow}$ & $0.31 \pm 0.20 \color{red}{\downarrow}$ \\
Ours (SCDP)          & $0.70 \pm 0.10 \color{green!60!black}{\uparrow}$ & $0.43 \pm 0.06 \color{red}{\downarrow}$ & $0.42 \pm 0.10$ & $0.80 \pm 0.12 \color{green!60!black}{\uparrow}$ \\
\cmidrule(lr){1-1}
Dataset             & $0.46 \pm 0.22$ & $0.64 \pm 0.25$ & $0.50 \pm 0.22$ & $0.64 \pm 0.24$ \\
\addlinespace 
\rowcolor{gray!15} \multicolumn{5}{l}{\textit{Navigation Evaluation}} \\
\addlinespace 
Diffusion Policy    & $0.42 \pm 0.03$ & $0.77 \pm 0.05$ & $0.58 \pm 0.16$ & $0.61 \pm 0.04$ \\
Legibility Diffuser & $0.85 \pm 0.04 \color{green!60!black}{\uparrow}$ & $0.02 \pm 0.03 \color{red}{\downarrow}$ & $1.06 \pm 0.28 \color{green!60!black}{\uparrow}$ & $-0.07 \pm 0.19 \color{red}{\downarrow}$ \\
Ours (SCDP)          & $0.64 \pm 0.02 \color{green!60!black}{\uparrow}$ & $0.47 \pm 0.05 \color{red}{\downarrow}$ & $0.31 \pm 0.01 \color{red}{\downarrow}$ & $0.85 \pm 0.08 \color{green!60!black}{\uparrow}$ \\
\cmidrule(lr){1-1}
Dataset             & $0.39 \pm 0.21$ & $0.70 \pm 0.25$ & $0.47 \pm 0.28$ & $0.64 \pm 0.26$ \\
\bottomrule
\end{tabularx}
\end{table}

\subsubsection{Performance Analysis}

Our results demonstrate that SCDP effectively balances trajectory legibility and path efficiency by adapting its behavior to the scene's spatial ambiguity.
\begin{itemize}
    \item[$\bullet$] Ambiguous Scenarios: In environments requiring higher clarity, the Legibility Diffuser baseline achieves the highest mean scores ($0.61$ in Block Reach and $0.62$ in Navigation). This is driven by its prioritization of detachment from negative goals, though it suffers from low efficiency. SCDP follows closely ($0.58$ and $0.59$ respectively), significantly outperforming the standard Diffusion Policy.
    \item[$\bullet$] Non-Ambiguous Scenarios: When the goal is clear, SCDP consistently outperforms all other methods, achieving the highest fused scores of ($0.74$ and $0.76$) in both tasks.
\end{itemize}
It is worth noting that across all experiments, the success rate for reaching the target goal remained above $0.98$ for all baselines and our proposed SCDP. This confirms that the style-conditioning modules modulate the trajectory path without compromising the underlying task performance of the base model.

\subsubsection{Discussion} While Legibility Diffuser can be tuned to maximize intent expression, our approach is limited to the data seen in the training set, meaning it is unlikely to generate trajectories that are more legible than those observed. Nonetheless, SCDP provides a data-driven alternative allowing for environment-awareness, that maintains high efficiency in clear scenarios, without explicit guidance tuning, offering a superior overall trade-off across diverse environmental contexts.

\section{Deployment}

\begin{figure}[tbp]
    \centering
    \includegraphics[width=1.0\linewidth]{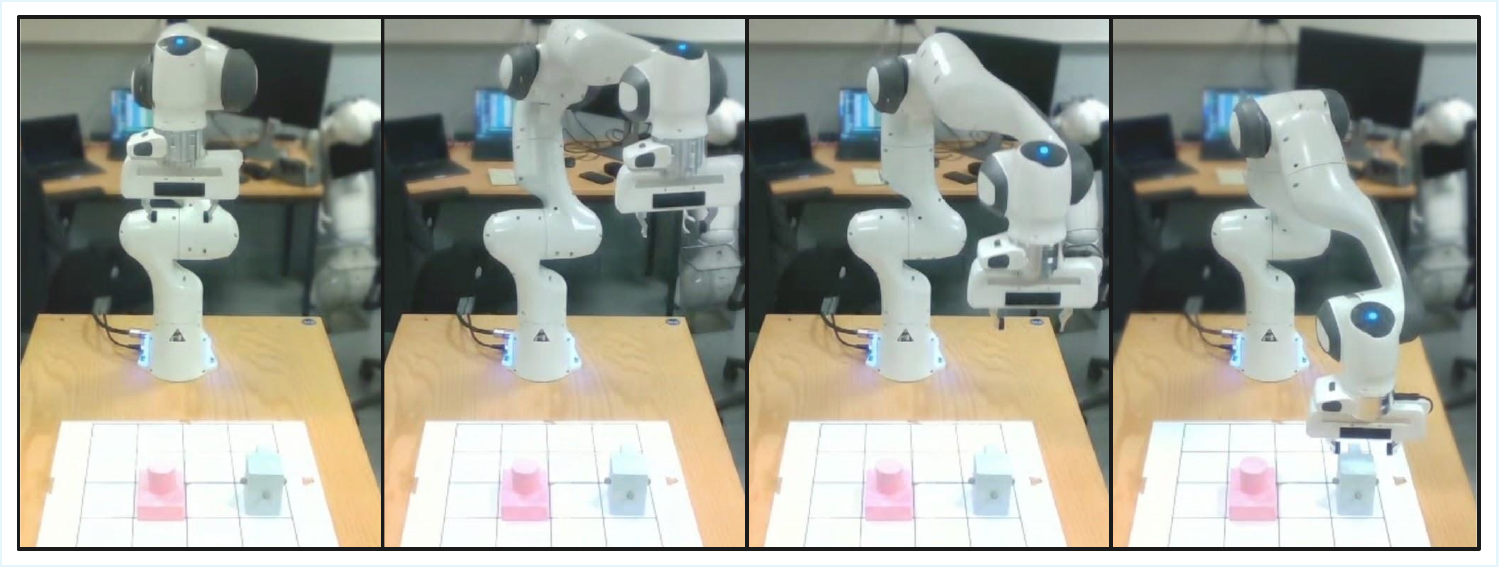}
    \caption{Real-world deployment of SCDP on a Franka Emika Panda robot for the Block Reach task. The image sequence (left to right) illustrates the model successfully generating an exaggerated, intent-expressive trajectory toward the target blue object to resolve spatial ambiguity relative to the distractor pink object.}
    \label{fig:franka}
\end{figure}

To evaluate the portability of SCDP beyond simulated environments, we deployed the model on a physical Franka Emika Panda robot to perform the Block Reach task, as shown in the figure \ref{fig:franka}.

\subsection{Perception}

To perform object detection, we use a fine-tuned YOLO \cite{yolo} model to detect target and distractor blocks from an Intel RealSense RGB-D's camera frame. The detected 2D bounding boxes are projected into the 3D space using the camera's depth map and extrinsic calibration with the robot's base frame. The obtained coordinates can then be used by the model.

\subsection{Sim-to-Real Transfer and Computational Efficiency}

We perform inference on a NVIDIA RTX A3000 GPU. The generated trajectories are interpolated to execute motion with a frequency of 1000 Hz. By using a state-based Diffusion Policy model, the model avoids any reality gap and mirrors simulation performance gains, without any specific calibration or fine-tuning on real demonstrations.
Our method achieves a mean total inference time of 5 seconds, which is similar to the base model's performance, due to our additional modules being lightweight and adding negligible latency.

\section{Conclusion}

We proposed an architecture that enables the specialization of a lightweight MLP from the initial model to learn a context-dependent trajectory style, that conditions a pretrained Diffusion Policy model to constrain the generation when needed, in our case toward legible and predictable motion. In addition to the baseline, these style-specific modules require smaller datasets for training and impose no overhead at inference time, offering a practical enhancement for diffusion-based policies.
While the focus of this work was primarily on legibility, we believe that the same learning method could be used to learn other trajectory-level concepts like safety, enabling style-conditioned motion generation through modular and reusable components.
Future work could aim to further develop ambiguity detection, as the current module can currently be considered as a geometrical proxy that could be swapped in a more complex ambiguity detector, and to conduct user studies to validate how human observers perceive and interpret these adaptive trajectories. Furthermore, future research could explore the framework's flexibility and scalability when faced with a higher number of potential goals.

\begin{credits}
\subsubsection{\ackname} This work used IDRIS HPC resources under the allocation 2025-[AD011017084] made by GENCI. It was funded by the French National Research Agency (ANR) under the OSTENSIVE project (ANR-24-CE33-6907-01) and the France 2030 program, reference ANR-23-PAVH-0005 (INNOVCARE Project). This project has received funding from the European Union's Horizon Europe Framework Programme under grant agreement No 101070596.

\subsubsection{\discintname}
The authors have no competing interests to declare that are relevant to the content of this article. 
\end{credits}
%
%
%
%

\end{document}